\documentclass[letterpaper, 10 pt, conference]{template/ieeeconf}  

\IEEEoverridecommandlockouts                              
\usepackage[utf8]{inputenc}
\usepackage[T1]{fontenc}

\usepackage{svg}    
\usepackage{amsmath} 
\usepackage{amssymb}  

\usepackage{mathtools}
\usepackage{subdepth} 
\usepackage{bm, upgreek} 

\usepackage{graphicx}

\usepackage{booktabs}

\usepackage{url}

\usepackage{placeins}
\FloatBarrier

\title{\LARGE \bf
InterLoc: LiDAR-based Intersection Localization\\ using Road Segmentation with Automated Evaluation Method
}

\author{Nguyen Hoang Khoi Tran, Julie Stephany Berrio, Mao Shan, Zhenxing Ming, and Stewart Worrall
\thanks{This work has been supported by the Vingroup Science and Technology Scholarship Program for Overseas Study for Master's and Doctoral Degrees, and the Australian Centre for Robotics (ACFR). The authors are with the ACFR at The University of Sydney (NSW, Australia). E-mails: \{n.tran, j.berrio, m.shan, d.ming, s.worrall\}@acfr.usyd.edu.au}
}

\begin{document}

\maketitle
\thispagestyle{empty}
\pagestyle{empty}

\begin{abstract}


Online localization of road intersections is beneficial for autonomous vehicle localization, mapping and motion planning. Intersections offer strong landmarks for correcting vehicle pose estimation, anchoring new sensor data in up-to-date maps, and guiding vehicle routing in road network graphs. 
Despite this importance, intersection localization has not been widely studied, with existing methods either ignoring the rich semantic information already computed onboard or relying on scarce, hand‑labeled intersection datasets. 
To close this gap, we present a novel LiDAR-based method for online vehicle-centric intersection localization. 
We detect the intersection candidates in a bird's eye view (BEV) representation formed by concatenating a sequence of semantic road scans. We then refine these candidates by analyzing the intersecting road branches and adjusting the intersection center point in a least-squares formulation. 
For evaluation, we introduce an automated pipeline that pairs localized intersection points with OpenStreetMap (OSM) intersection nodes using precise GNSS/INS ground‑truth poses. 
Experiments on the SemanticKITTI dataset show that our method outperforms the latest learning-based baseline in accuracy and reliability. 
Sensitivity tests demonstrate the method's robustness to challenging segmentation errors, highlighting its applicability in the real world.

\end{abstract}

\section{Introduction}


Autonomous vehicles continuously reason about the topology and geometry of the road network that surrounds them. Among the various features, intersections are especially valuable: they provide reliable landmarks that can suppress accumulated odometry drift under GNSS degradation, anchor road network graphs for mapping and routing, and impose constraints on motion planning for safe, lawful maneuvering. Although visual cues have been widely exploited for intersection detection, camera‑based methods remain vulnerable to adverse illumination and weather conditions. LiDAR offers structural information that is invariant to lighting, yet the literature on LiDAR‑only intersection localization is relatively sparse, and existing approaches tend to disregard the rich semantic information now routinely produced by modern segmentation networks. 

\begin{figure} [t]
    \centering
    \includegraphics[width=1\linewidth]{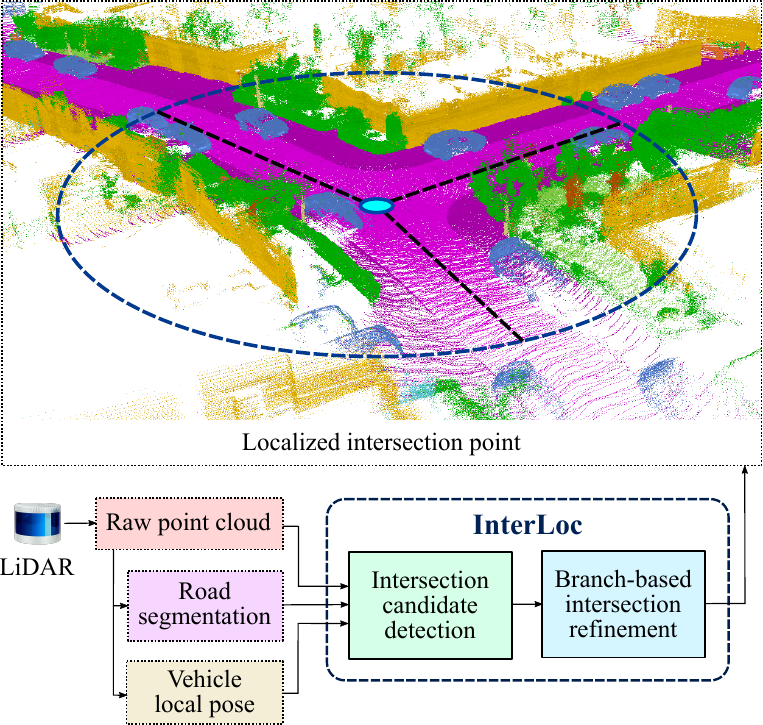}
    \caption{Intersection localization from LiDAR data. Top: An intersection point is localized on the concatenation of semantic point clouds. Bottom: Road segmentation and vehicle local pose are computed from the raw point cloud. Our algorithm utilizes these three pieces of information to detect and localize intersection points through a coarse-to-fine process.
   \vspace{-5mm} }
    \label{fig:overview}
\end{figure}


Our work addresses two coupled challenges. The first is the accurate localization of intersection center points using only vehicle‑mounted LiDAR scans. The second is the quantitative evaluation of localization performance without relying on hand‑annotated intersection labels, which are time‑consuming to produce and absent from most public datasets. 


To solve the first challenge, we leverage off‑the‑shelf deep‑learning semantic segmentation to isolate road points, concatenate multiple keyframes using locally estimated pose to obtain a balanced panoramic scan, project the resulting scan onto a bird's eye view (BEV) grid, and detect candidate intersections as high‑saliency corners on the skeletonized road centerline. The candidate points are then verified and refined by analyzing the topology of outgoing branches: points whose neighborhood contains at least three branches are retained, and their positions are adjusted to minimize perpendicular residuals to straight‑line branch approximations.


The second challenge is met by an automated benchmarking pipeline that aligns LiDAR-based detections with OpenStreetMap (OSM) road graphs. The precise GNSS/INS poses available in many datasets make it possible to georeference each detected intersection point and associate it with the nearest OSM intersection node within a confidence radius. This process provides ground-truth correspondence for computing intersection localization error and identifying false positive (FP) and false negative (FN) detections, enabling the standard metrics of precision, recall and average center error to be measured at scale without manual intervention.

Experiments on the SemanticKITTI dataset validate the effectiveness of the proposed framework. Across eight different driving sequences, the method achieves sub-two-meter localization accuracy and exceeds a recent learning-based baseline in both precision and recall. Moreover, sensitivity analyses in which segmentation masks are corrupted with synthetic noise reveal performance degradation: localization errors remains within 0.4 m of the benchmark model when FP and FN noises are increased by factors of 3.7 and 7, respectively. These findings highlight the practicality of integrating the developed method into existing autonomous driving systems that already compute semantic masks for other purposes.

In summary, the main contributions of this paper are:
\begin{enumerate}
    \item We propose a novel LiDAR-based intersection localization method using road segmentation and vehicle local pose. To the best of our knowledge, our method is the first one to make use of road segmentation as input for this task.
    \item We propose an automated evaluation process for intersection localization using georeferenced road graph. This bypasses the dependency on manual intersection labels, which are currently absent in public datasets.
    \item We conduct comprehensive performance analysis of the method. We compare the accuracy of the method with the state-of-the-art and analyze its robustness under challenging segmentation error conditions, which is essential for real-world applications. 
\end{enumerate}



\section{Related Work}\label{literature}

Intersections are part of road network graphs. Prior road graphs can be generated through manual annotation \cite{OpenStreetMap}, GNSS traces \cite{wang_automatic_2017}, aerial imagery \cite{hetang_segment_2024}, and multimodal fusion \cite{li_guided_2021}. To use these prior road graphs during online operations, the vehicle's position must be referenced in the global coordinate frame, typically via GNSS. However, GNSS is unreliable in urban canyons and areas with overhead occlusions. Additionally, prior road maps can be outdated over time and requires frequent updates \cite{9484742}. 
To address these issues, intersection detection methods using on-vehicle perception have been developed, with cameras and LiDAR being the most commonly used sensors.

\subsection{Camera-based Approach}
Camera-based methods focus on detecting and classifying intersections at the frame level. Early methods used semantic segmentation extracted by conditional random fields \cite{lafferty2001conditional} to classify multiple road geometries \cite{ess2009segmentation}. Spatial data and odometry from stereo cameras were then incorporated to improve classification performance \cite{ballardini_online_2017}. With the development of deep learning, convolution neural networks were combined with long short-term memory \cite{hochreiter1997long} to leverage spatiotemporal understanding from a video sequence \cite{bhatt_have_2017}. To enhance robustness, the author of \cite{koji_deep_2019} merged sequential and single-frame approaches into a unified network. The single-frame approach was further developed using very deep networks with the ResNet architecture \cite{he2016deep} for standard and panoramic images  \cite{astrid_for_2020, sugimoto_intersection_2022}. Their reliability was enhanced by applying the teacher/student training paradigm \cite{ballardini_model_2021}. Despite promising results, existing camera-based methods have not yet demonstrated the ability to localize intersection points in spatial coordinates. In addition, the quality of camera data inherently depends on lighting and weather conditions.

\subsection{LiDAR-based Approach}
Compared to cameras, LiDAR has the advantage of being independent of lighting conditions and relatively resistant to inclement weather. Early LiDAR-based intersection classifier employed beam models and peak-finding algorithms to identify the number of branches of the road ahead \cite{chen_lidar-based_2011,zhang_3d_2015}. Classification accuracy was then improved by applying traditional machine learning techniques \cite{zhu_3d_2012, hata_road_2014, habermann_road_2016}. Deep learning-based methods were also studied. \cite{baumann_classifying_2018} proposed a transfer learning technique to deal with small size datasets. \cite{yan2020lidar} designed a multi-task network that performs road geometry classification simultaneously with road segmentation and height estimation. 
Although LiDAR data contains precise spatial information, there is little research on its use for intersection localization. In \cite{wang_3d-lidar_2017}, the intersection point was estimated from a multi-frame LiDAR point cloud with the help of wheel odometry, GNSS and OSM priors. Recently, a deep learning network was developed to classify and localize intersections from a single LiDAR frame \cite{li_intersection_2024}. These methods have overlooked the available semantic information that can be effectively extracted from point clouds using recent segmentation techniques \cite{milioto_rangenet_2019}. In addition, there is a lack of public datasets that contain ground-truth intersection positions in the vehicle coordinate frame. This hinders the training and evaluation of methods. 

To close these gaps, we propose a novel LiDAR-based intersection detection and localization method that uses road segmentation information as input. The method is modular so that it can be easily coupled with off-the-shelf road segmentation frameworks. To address the lack of ground truth, we design a non-learning algorithm based on morphology, salient point detection, and branch geometry analysis. In addition, we propose an automated evaluation method that utilizes the intersection nodes from georeferenced road maps. We use the vehicle ground truth poses available in many public datasets to register the estimated intersection positions to the corresponding global coordinates.

\section{LiDAR-based Intersection Localization using Road Segmentation}\label{model}

This section addresses intersection localization using LiDAR data. Fig. \ref{fig:overview} shows an overview of our method. First, we detect intersection candidate points from LiDAR scans with the assistance of road segmentation and vehicle local pose. Then, we classify and refine the intersection by analyzing the connected branches. The two stages are presented in Sections \ref{sec:lid_fext} and \ref{sec:its_refine}.

\subsection{Definitions}

We define \((W)\) as the world coordinate frame fixed at the initial pose of the LiDAR. \((L_k)\) represents the LiDAR coordinate frame at the time step \(k\). 
The pose of \((L_k)\) with respect to \((W)\) is defined as the transformation matrix  \( \bm{T}_{WL_k} = (\bm{R}_{WL_k}, {_W}{\bm{p}}_{L_k}) \in SE(3) \), where \( \bm{R} \in SO(3)\) is the rotation matrix and \(\bm{p} \in \mathbb{R}^3 \) is the position vector. We denote the LiDAR point cloud at time \(k\) expressed in $(L_k)$ as \(_{L_k}\mathcal{P}_k = \{ _{L_k}\bm{p}_{k,i}\} \), where \(i\) is the point index. The corresponding semantic labels of the point cloud are represented as \(\mathcal{S}_k = \{s_{k,i}\}\). For each time step, we know the raw point cloud \(_{L_k}\mathcal{P}_k\), semantic labels \(\mathcal{S}_k\), and vehicle local pose \(\bm{T}_{WL_k}\).

\begin{figure} [t]
    \centering
    \includegraphics[width=0.75\linewidth]{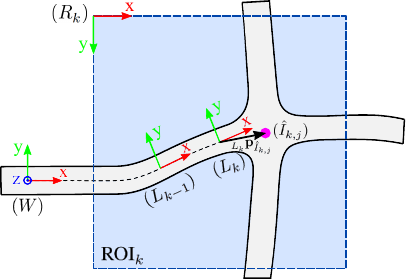}
    \caption{Illustration of coordinate frames and notation used in this work. We sample a region of interest (blue square) around the current LiDAR frame \((L_k)\). From this, we detect the intersection point (magenta) and estimate its LiDAR-centric position (black arrow).
\vspace{-5mm}     }
    \label{fig:coord_frames}
\end{figure}

We define the coordinate frame of the candidate and refined intersection point as \(({I}_{k,j})\) and \((\hat{I}_{k,j})\), respectively, where \(j\) is the intersection index at time step \(k\). We denote \((R_k)\) as the region-of-interest (ROI) coordinate frame at time step \(k\). 
To represent a 2D pose on the x-y plane, we denote the transformation matrix \(\mathbf{T}=(\mathbf{R}, \mathbf{p}) \in SE(2)\), where \(\mathbf{R} \in SO(2)\) is the rotation around the z axis and \(\mathbf{p} \in \mathbb{R}^2 \) is the x-y position. Our algorithm estimates the 2D LiDAR-centric position of each refined intersection point, denoted as \( {_{L_k}}\mathbf{p}_{\hat{I}_{k,j}} \). To perform operations in the discrete image space, we also denote \( \bm{\uprho}\in \mathbb{R}^2\) as a 2D position in the image coordinate frame. Fig. \ref{fig:coord_frames} visualizes the defined coordinate frames and notation.

\subsection{Intersection Candidate Detection} \label{sec:lid_fext}

\begin{figure} [t]
    \centering
    \includegraphics[width=1\linewidth]{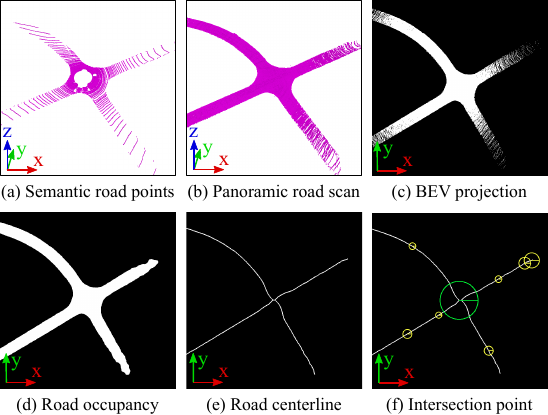}
    \caption{Step-by-step visualization of the intersection candidate detection process. (a) We segment 3D road points from each input scan using semantic information. (b) We concatenate multiple keyframes using vehicle local poses to construct a panoramic road scan. (c) We project this scan into a 2D BEV binary image. (d) We infer the road occupancy and (e) extract the road centerline on the BEV plane. (f) Finally, we identify the intersection candidate point (green circle) based on its superior Harris score (shown as the circle radii) compared to the other corner points (yellow circles).
\vspace{-5mm}     }
    \label{fig:lid_fext}
\end{figure}

Fig. \ref{fig:lid_fext} illustrates the process of detecting intersection candidates from LiDAR data. The process involves six steps: segmentation of the road points, concatenation of the keyframes, projection to BEV, inference of road occupancy, extraction of the road centerline, and detection of intersection points. Each step is detailed in the following subsections.

\subsubsection{Road point segmentation}
Deep learning-based LiDAR semantic segmentation methods typically achieve high accuracy for the road class. To leverage that accuracy, we use the labels provided by such methods to segment road points from the raw point cloud. The segmented road scan at time step \(k\), denoted as \({ _{L_k}}\mathcal{P}^{\mathrm{road}}_k\), is the set of points in \(_{L_k}\mathcal{P}_k\) that are labeled as \textit{road} in \(\mathcal{S}_k\):
\begin{equation}
    { _{L_k}}\mathcal{P}^{\mathrm{road}}_k=\{ _{L_k}\bm{p}_{k,i} \mid s_{k,i}= \textit{road} \}. 
\end{equation}

\subsubsection{Keyframe concatenation}
Fig. \ref{fig:lid_fext}(a) indicates that a single road scan has an uneven point density and contains occluded areas. This could cause inaccuracies in subsequent processing. To overcome this, we concatenate multiple road scans from consecutive \textit{keyframes} using LiDAR poses. We follow an empirical approach to keyframe selection as in \cite{shan_lio-sam_2020}. We select a LiDAR frame as a keyframe when its pose change compared to the previous keyframe exceeds a user-defined threshold. This approach eliminates redundant frames during slow vehicle movement and diversifies the scanning viewpoints. 
We denote the relative position and angle thresholds for keyframe selection as $\delta^\mathrm{p}$ and $\delta^\mathrm{a}$. We define the concatenated road scan at time step \(k\) as: 
\begin{equation}
    {_W}\mathcal{P}^\mathrm{acc}_k=\{ {_W}\mathcal{P}^{\mathrm{road}}_{k-n},..., {_W}\mathcal{P}^{\mathrm{road}}_{k+n}\}, 
\end{equation}
where \(n\) is the number of keyframes before and after the current keyframe. The component point clouds are transformed into \((W)\) using their corresponding LiDAR poses \(  \{ \bm{T}_{WL_{k-n}},..., \bm{T}_{WL_{k+n}} \} \).  Fig. \ref{fig:lid_fext}(b) shows that \({_W}\mathcal{P}^\mathrm{acc}_k\) achieves a more evenly distributed point density and complete coverage of the local road region.

\subsubsection{BEV projection}
We project the concatenated point cloud onto the x-y plane of \((W)\). Since the roads are flat in the local area and have a slight slope toward the horizontal plane, this projection compresses the data from 3D to 2D while preserving the main shape of the road. We discretize the projected scan into a binary BEV image. This further reduces the data size and enables the use of efficient binary image processing techniques.
Specifically, we sample the projection of \({_W}\mathcal{P}^\mathrm{acc}_k\) within a region of interest (ROI) of size \(S \times S\), centered at \((L_k)\). This ROI is aligned with the x and y axes of \((W)\). We divide the ROI into equal cells \(c_{uv}\) with resolution \(r\), where \(u, v\in \{0,1,..., \lfloor \frac{S}{r} \rfloor\}\). Let \(\mathbf{p}^{\mathrm{acc}}_i\) denote a point in the projected \({_W}\mathcal{P}^\mathrm{acc}_k\). If at least \(m\) points of \(\mathbf{p}^{\mathrm{acc}}_i\) fall within a cell \(c_{uv}\), we set the corresponding pixel \(b_{uv}\) in the BEV image. The BEV image is defined as: 
\begin{equation}
     \mathbf{I}^{\mathrm{bev}}_k = \{b_{uv}\}, \,\,\,
     b_{uv} =
        \begin{cases}
            1 &  |\{ \mathbf{p}^\mathrm{acc}_i \in c_{uv}\}| \ge m \\
            0 & \text{otherwise}.
        \end{cases}
\end{equation}
The threshold \(m \) suppresses low-density cells, which often include outliers and false-positive segmented points. Fig. \ref{fig:lid_fext}(c) depicts the resulting \(\mathbf{I}^{\mathrm{bev}}_k\).

\subsubsection{Road occupancy inference}
We construct a continuous object in the BEV plane that represents the area of the road, called the \textit{road occupancy}. First, we apply a morphological closing \cite{talbot_mathematical_2010} to the binary BEV image. This operation fills the spaces between positive pixels to create continuous objects in the image. Subsequently, we apply a morphological opening \cite{talbot_mathematical_2010} to the resulting image. This removes small objects formed by isolated points and smooths out the main object. The road occupancy image is defined as:
\begin{equation}
     \mathbf{I}^{\mathrm{occ}}_k = \mathbf{I}^{\mathrm{bev}}_k \bullet B^\mathrm{c} \circ B^\mathrm{o}
\end{equation}
where \(\bullet\) and \(\circ\) denote morphological closing and opening operators, respectively. \(B^\mathrm{c}\) and \(B^\mathrm{o}\) are the structuring elements applied to the operators. We use circular elements to create a smooth object edge. Fig. \ref{fig:lid_fext}(d) illustrates the computed \(\mathbf{I}^{\mathrm{occ}}_k\).

\subsubsection{Road centerline extraction}
The road centerline is the medial axis of the road occupancy object. It preserves the topology and shape of the road in a one-pixel-wide object. We skeletonize the road occupancy image to obtain the road centerline image as:
\begin{equation}
     \mathbf{I}^{\mathrm{cen}}_k = \mathcal{S}(\mathbf{I}^{\mathrm{occ}}_k )
\end{equation}
where \(\mathcal{S}(.)\) denotes the Zhang-Suen thinning algorithm \cite{zhang_fast_1984}. 

\subsubsection{Intersection candidate point detection}
Fig. \ref{fig:lid_fext}(e) shows that the intersection candidate point corresponds to the high-gradient position in the road centerline image. Therefore, we extract the positions of intersection candidates in the image coordinates as:
\begin{equation}
     \{_{R_k}\bm{\uprho}_{I_{k,j}}\} =\mathcal{C}(\mathbf{I}^{\mathrm{cen}}_k)
\end{equation}
where \(\mathcal{C}(.)\) is the Harris corner detector \cite{harris1988combined}. This method is reliable on \(\mathbf{I}^{\mathrm{cen}}_k\) because it is a synthetic image with low noise and high contrast. Fig. \ref{fig:lid_fext}(f) visualizes an extracted point.

\subsection{Branch-based Intersection Refinement} \label{sec:its_refine}

\begin{figure} 
    \centering
    \includegraphics[width=1\linewidth]{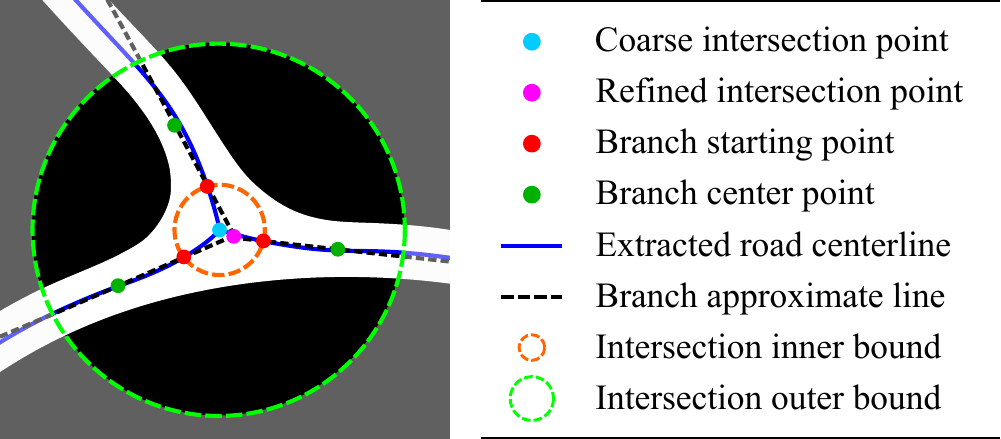}
    \caption{Visualization of the intersection refinement method. We approximate the branches around the intersection candidate point (cyan) as straight lines (dashed black). We compute the refined intersection point (magenta) that achieves the least squares of the perpendicular distances to the approximate lines.
    \vspace{-5mm} }
    \label{fig:its_refine}
\end{figure}

The intersection candidate points detected in the first stage are coarse and pose two problems. First, the area around the candidate point does not always contain enough valid branches to be called an intersection. Second, the position of the candidate point may deviate from the actual intersection point. These problems are caused by road segmentation errors and the inherent curvature of the road boundary in the intersection area. In this stage, we propose a branch-based method to classify and refine the candidate point. We segment the intersecting branches around the point, classify whether the area is an intersection, and correct the position of the intersection point. Fig. \ref{fig:its_refine} illustrates the method.

\subsubsection{Branch segmentation}
Around each intersection candidate point, we define two areas in the image of the road centerline: an inner disk $ \mathcal{A}^\mathrm{i}$ and an outer annulus $ \mathcal{A}^\mathrm{o}$. These two regions are adjacent and are centered at the candidate point. We formulate them as:
\begin{align}
        \mathcal{A}^\mathrm{i} &=  \mathbf{I}^{\mathrm{cen}}_k \cap \{\| \bm{\uprho} - {_{R_k}}\bm{\uprho}_{I_{k,j}} \| < a^\mathrm{i}\} \\
    \mathcal{A}^\mathrm{o} &= \mathbf{I}^{\mathrm{cen}}_k \cap \{a^\mathrm{i} < \| \bm{\uprho} - {_{R_k}}\bm{\uprho}_{I_{k,j}} \| < a^\mathrm{o} \}
\end{align}
where $a^\mathrm{i}$ and $a^\mathrm{o}$ are the radii of the inner and outer boundary circles, respectively. If there are other candidate points in the inner disk, we merge them and the point under consideration into one point at the average position, and re-establish the two regions around the merged point. 
In the outer annulus, we segment the road branches as connected positive pixel objects. A valid branch starts from the inner circle boundary and ends before any other intersection candidate points. We approximate each branch with a radial line \(\beta_b\), where $b$ is the branch index. This line passes through the branch starting point $\bm{\uprho}^\mathrm{s}_b$ that lies on the inner circle boundary. To best fit the branch, this line also passes through the branch center point \(\bm{\uprho}^\mathrm{c}_b\), which is the mean position of all pixels in the branch. In sum, we define the branch approximate line as:
\begin{equation}
     \beta_b  = \{(1-t)\bm{\uprho}^\mathrm{s}_b + t\bm{\uprho}^\mathrm{c}_b  \mid t \in \mathbb{R}\}, \;\;  \{\bm{\uprho}^\mathrm{s}_b , \bm{\uprho}^\mathrm{c}_b\} = \mathcal{B}(\mathcal{A}^\mathrm{o})
\label{eqn:its_branch}
\end{equation}
where $\mathcal{B}(.)$ is the function that extracts the starting and center points of the branches as described. 

\subsubsection{Intersection classification and correction}
If the number of segmented branches $B$ is greater than or equal to three, we classify the candidate as an intersection and vice versa. We compute the refined intersection point, which achieves the least squares of the perpendicular distances \(d_\perp\) to the approximate branch lines as:
\begin{equation}
    {_{R_k}}\bm{\uprho}_{\hat{I}_{k,j}} = \underset{ \bm{\uprho} \in \mathcal{A}^\mathrm{i} }{ \arg\!\min}  \left\{ \sum_{b=1}^{B}{d_\perp(\bm{\uprho}, \beta_b)^2} \right\}
      \label{eqn:alter_its}
\end{equation}
This equation has a unique solution when there are at least two non-parallel branch lines, which is usually the case at intersections. Finally, we convert the refined point into the LiDAR coordinate frame in the continuous space as: 
\begin{equation}
      {}_{L_k}\mathbf{p}_{\hat{I}_{k,j}} =r\mathbf{T}_{{L_k}{R_k}}{_{R_k}}\bm{\uprho}_{\hat{I}_{k,j}}
      \label{eqn:lidcen_its}
\end{equation}
where $\mathbf{T}_{{L_k}{R_k}} = (\mathbf{R}_{R_kW}\mathbf{R}_{WL_k},[\frac{S}{2}, \frac{S}{2}])^{-1}$. $\mathbf{R}_{R_kW}$ is the rotation of $\pi$ radians around the $x$-axis. $\mathbf{R}_{WL_k}$ is given by the LiDAR pose.

\section{Automated Evaluation Method based on Georeferenced Road Maps}\label{eval}

\begin{figure} 
    \centering
    \includegraphics[width=1\linewidth]{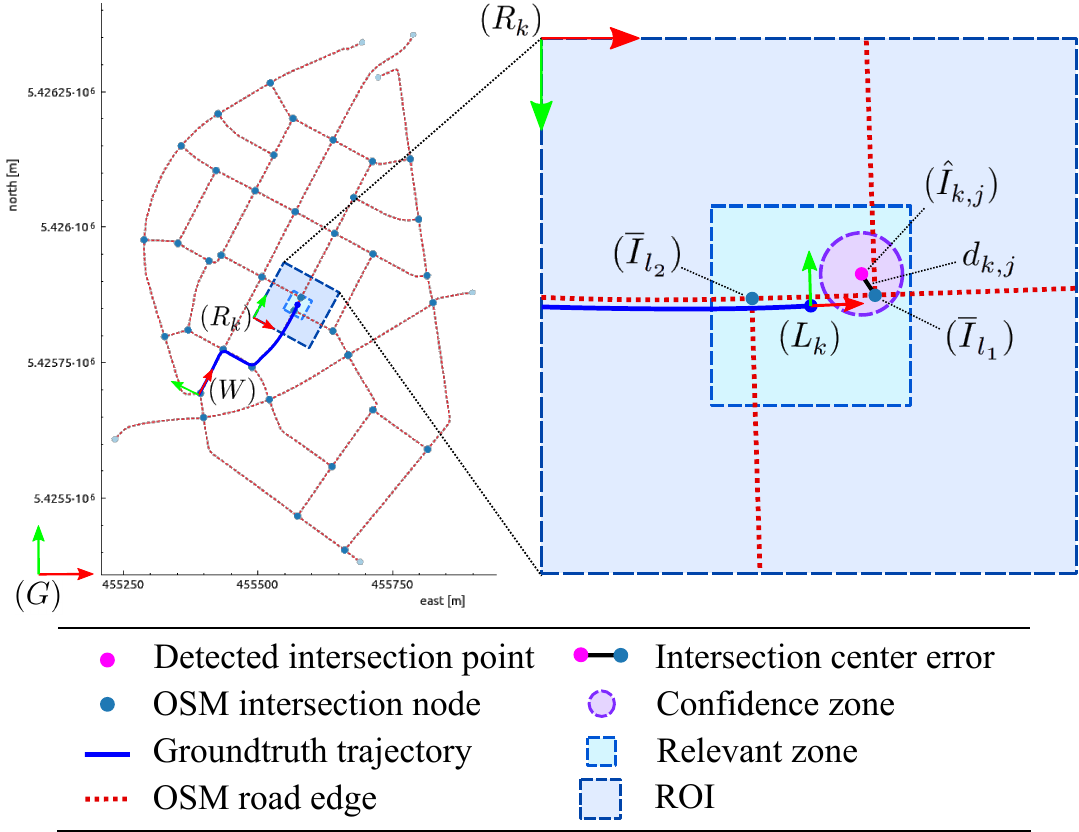}
    \caption{Evaluation method for intersection detection using the OSM road map. The detected intersection point $(\hat{I}_{k,j})$ is matched to the closest OSM intersection node $(\overline{I}_{l_1})$ as its ground truth. In this example, the detection is a TP because the ground truth point lies within the confidence zone of the detected point. The other OSM intersection node $(\overline{I}_{l_2})$ is a FN because it lies within the relevant zone of the current LiDAR frame $(L_k)$ and does not match any detected points.
    \vspace{-5mm} }
    \label{fig:eval_osm}
\end{figure}

In this section, we present a method for automatically evaluating the localized intersections using the OSM road map. We first find the ground-truth intersection nodes from the OSM road map that match the detected intersections. We then calculate evaluation metrics based on the relative positions of the detected intersection, the ground-truth intersection, and the current LiDAR frame. Fig. \ref{fig:eval_osm} visualizes the method.

\subsection{Intersection Ground Truth Automated Generation}

The ground truth pose of the vehicle in the georeferenced coordinate system $(G)$ is denoted as ${\overline{\mathbf{T}}}_{GL_k}$. This information is provided by GNSS/INS sensors and is often available in datasets for autonomous driving. We use it to convert the position of each detected intersection point to $(G)$ as follows:
\begin{equation} 
        {_G}{\mathbf{p}}_{\hat{I}_{k,j}} = {\overline{\mathbf{T}}}_{GL_k} {}_{L_k}{\mathbf{p}}_{\hat{I}_{k,j}}
    \label{eqn:dtc_its}
\end{equation}
We extract from the OSM a georeferenced road map of the area around the vehicle's position. This map includes intersection nodes that we use as ground truth to evaluate the detected intersection points. 
We match these OSM intersection nodes with the detected intersection points. From Eq. \ref{eqn:dtc_its}, we calculate the detected intersection points in the same coordinate system $(G)$ as the OSM intersection nodes. We search for OSM intersection nodes $(\overline{I}_l)$ that are in the ROI around $(L_k)$, where $l$ is the node index. We denote the set of positions of these nodes as $\mathcal{O}_k=\{{_G}\mathbf{p}_{\overline{I}_{l}}\}$. Then, we match each detected intersection point $(\hat{I}_{k,j})$ with the closest point in $\mathcal{O}_k$ by Euclidean distance as:
\begin{equation} 
    {_G}\mathbf{p}_{\overline{I}_{k,j}} = \underset{ \mathbf{p} \in\mathcal{O}_k }{ \arg\!\min} \Vert \mathbf{p} - {_G}\mathbf{p}_{\hat{I}_{k,j}} \Vert
    \label{eqn:gth_its}
\end{equation}
where $(\overline{I}_{k,j})$ stands for the OSM intersection node that matches the detected intersection under consideration. Some detected intersection nodes may not have a matching OSM intersection node, and vice versa. All such cases are accounted for in the evaluation, using metrics defined in Section \ref{sec:metrics}.

\subsection{Evaluation Metrics}
\label{sec:metrics}
\subsubsection{Average Center Error (ACE)} 
This metric measures the localization accuracy of detected intersection points. It was first proposed in \cite{li_intersection_2024}. We define the center error as the Euclidean distance between each detected intersection and its matched ground truth intersection. Then, we calculate the average of center errors of all matched pairs throughout the dataset as:
\begin{equation}
ACE =\frac{\sum_{k,j}{d_{k,j}}}{\sum_{k}{q_{k}}}, \;\;   d_{k,j} = \Vert {}_G \mathbf{p}_{\hat{I}_{k,j}} - {}_G \mathbf{p}_{\bar{I}_{k,j}} \Vert
\end{equation}
where $q_k$ is the number of matched pairs at time step $k$.

\subsubsection{Precision and Recall}
We measure the ratio if a detected intersection is true (precision), and if a true intersection is detected (recall). A true positive (TP) is counted when a detected intersection has a matched ground truth intersection, and its center error is smaller than a confidence threshold $D$. This means that the ground-truth intersection lies in a confidence circle around the detected point. A FP is counted otherwise.
\begin{align}
    TP &: \exists {_G}\mathbf{p}_{\overline{I}_{k,j}} \;\; \text{s.t}. \;\; d_{k,j} < D \\
    FP &: \text{otherwise}
\end{align}
We define the relevant zone as the area in which the detection method should identify any intersection points, if they exist. A FN is counted when a ground truth intersection lies in the relevant zone $\mathcal{Z}^\mathrm{r}_k$ around the current position of the LiDAR but does not have a matching detection. In our case, this zone is a square with side length $S - 2a^\mathrm{o}$ and is centered within the ROI. 
\begin{align}
    FN &:  \exists{}{_G}\mathbf{p}_{\overline{I}_{l}} \in \mathcal{Z}_k^\mathrm{r} \;\; \text{s.t.} \;\; {_G}\mathbf{p}_{\hat{I}_{l}} = \varnothing
\end{align}
Finally, we calculate the precision (Pre) and recall (Rec) as:
\begin{equation} 
    \text{Pre} = \frac{TP}{TP+FP},\quad
    \text{Rec} = \frac{TP}{TP+FN}
    \label{eqn:pre_rec}
\end{equation} 
\vspace{-2mm}

\section{Experimental Results}\label{exp}

In this section, we present the evaluation results of our intersection localization method. We explain the experimental setup, show the performance compared to the state-of-the-art, and demonstrate the robustness with different levels of road segmentation errors.

\begin{table}[b]
    \centering
    \vspace{-2mm}
    \caption{Parameters of our intersection detection method}
    \begin{tabular}{l  |l l l} 
        \toprule
        Module & \multicolumn{3}{c}{Parameters} \\
        \midrule
        Keyframes & $\delta^\mathrm{p} = 2$ m & $\delta^\mathrm{a}\ = 5^\circ$ & $n = 20$\\
        BEV image & $S = 120$ m & $r = 0.16$ m & $m = 5$\\
        Refinement & $a^\mathrm{i} = 10$ m & $a^\mathrm{o} = 40$ m & \\
        \bottomrule
    \end{tabular}
    \label{tab:params}
\end{table}

\begin{figure*} [t]
    \centering
    \includegraphics[width=\linewidth]{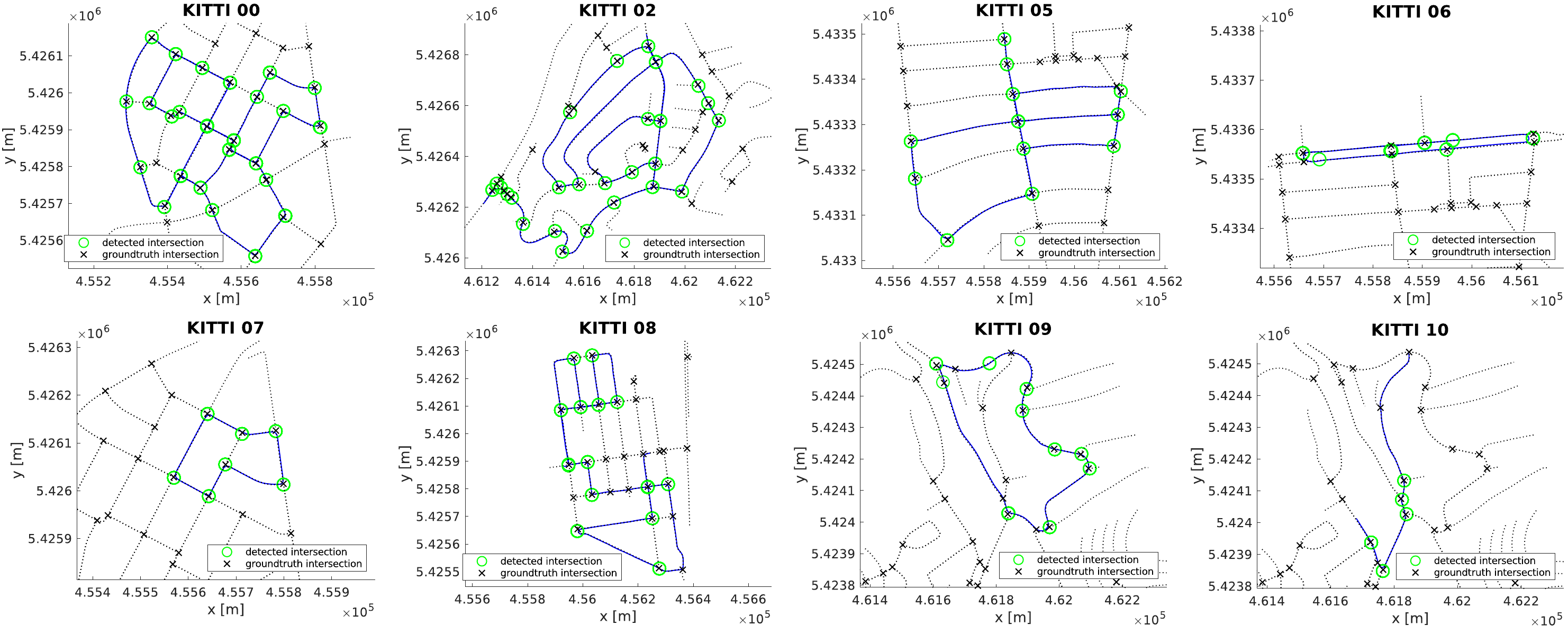}
    \caption{Georeferenced positions of detected intersections on the SemanticKITTI dataset sequences. Green: detected intersections. Black crosses: ground truth intersections. Black dashed: ground truth roads. Blue: ground truth trajectory. 
    \vspace{-5mm} }
    \label{fig:det_res}
\end{figure*}

\begin{table}[t]
    \centering
    \caption{Performance on SemanticKITTI sequences}
    \resizebox{\columnwidth}{!} {%
        \begin{tabular}{l | c c c c c c c c | c} 
            \toprule
            Sequence & 00 & 02 & 05 & 06 & 07 & 08 & 09 & 10 & Average\\
            \midrule 
            ACE [m] & 1.28 & 2.76 & 0.62 & 4.21 & 1.25 & 1.66 & 3.16 & 3.01 & 1.92 \\
            Pre@5 [\%]& 97.23 & 81.47 & 100.0& 44.72 & 100.0& 91.86 & 78.57 & 100.0& 89.48 \\
            Rec@5 [\%]& 94.15 & 72.95 & 91.87 & 25.90 & 77.30 & 96.86 & 52.84 & 67.57 & 76.74 \\        
            \bottomrule
        \end{tabular}
    }
    \vspace{-4mm}
    \label{tab:det_eval}
\end{table}


\subsection{Experimental Setup}

\subsubsection{Dataset}
We tested the method on the SemanticKITTI dataset \cite{behley_semantickitti_2019}. We used 8 of the 11 data sequences that provide ground-truth semantic labels, namely: 00, 02, 05, 06, 07, 08, 09, 10. The remaining three sequences are short and contain few intersections. The test sequences have a total of 21028 LiDAR scans and a traveled distance of 18.76 km. Driving scenes encompass residential areas, city streets, and highways. The LiDAR scans have 64 beams and were recorded at 10 Hz. The dataset provides locally-referenced LiDAR poses estimated by the SuMa algorithm \cite{behley_efficient_2018}. We ran the RangeNet++ \cite{milioto_rangenet_2019} algorithm on the LiDAR scans to generate semantic labels as input for our algorithm. We also use the hand-annotated ground-truth semantic labels provided in SemanticKITTI to test the algorithm's robustness under different segmentation errors.

We obtained the georeferenced ground truth poses from the original KITTI dataset \cite{geiger_vision_2013}. We transformed the provided GNSS/INS poses into LiDAR poses using calibration information. We sourced georeferenced road maps from OSM \cite{OpenStreetMap} and transformed the data into a road graph format using \emph{osm2pgrouting}. Then, we extracted the graph nodes with at least three connected edges to serve as ground truth intersections.

\subsubsection{Parameter settings}

Table \ref{tab:params} shows the parameters of our method. These parameters are kept constant in all experiments. We select the parameters through experimentation so that the algorithm achieves high accuracy and runs efficiently in real time. We choose $\delta^\mathrm{p}$ and $\delta^\mathrm{a}$ to balance detail and coverage of the concatenated scan. Similarly, $n$ and $S$ balance between scene completeness and computational cost. $m$ is chosen to filter out outliers but not to eliminate sparsely distributed true points. $a^\mathrm{i}$ and $a^\mathrm{o}$ are chosen to fit the size and curvature of most intersections in the dataset. With limited computational resources, one can change the resolution $r$ to $0.25$ m or $0.5$ m to speed up the processing with a trade-off in accuracy. 

\subsection{Performance Analysis and Comparison} \label{sec:performance}

\subsubsection{Performance on SemanticKITTI dataset}

Fig. \ref{fig:det_res} shows the qualitative results of our method in all test sequences. We can see that the sequences go through a wide variety of intersection shapes. Under that condition, most of the estimated intersections coincide with the ground truth intersections. Table \ref{tab:det_eval} shows the quantitative results. On average, the ACE is less than $2$ m, the precision and recall at the tolerance $D=5$ m are $89$\% and $77$\%, respectively. These results indicate that our method is reliable and accurate. Some specific cases cause performance degradation, which will be presented in Section \ref{sec:exp_limit}.

\subsubsection{Comparison with MMInsectDet}

\begin{table} [t]
    \centering
    \caption{Performance comparison with MMInsectDet}
    \begin{tabular}{l | c c c} 
        \toprule
        Method & ACE [m]& Pre@6.9 [\%]& Rec@13.3 [\%]\\
        \midrule
        MMInsectDet \cite{li_intersection_2024} & 4.25 & 89.23 & 83.10\\
        InterLoc (ours)& \textbf{1.92}& \textbf{94.38}& \textbf{84.28}\\
        \bottomrule
    \end{tabular}
    \vspace{-5mm}
    \label{tab:det_cmp}
\end{table}

We compare our method with MMInsectDet \cite{li_intersection_2024}, a state-of-the-art LiDAR-based intersection detection method. We used the results in the paper for comparison since their source code and intersection dataset have not been published. 
Their evaluation dataset and ours both originate from the KITTI dataset \cite{geiger_vision_2013}. They also used ACE, precision, and recall as evaluation metrics. To our knowledge, there is no other method similar enough to compare with ours. The TP and FP distinction in MMInsectDet used the CEIOU quantity, which combines $80$\% of the center error and $20$\% of the intersection over union (IoU). We assume an average IoU value of $0.5$ to convert the CEIOU to the distance threshold $D$ using the formula in the paper. Accordingly, CEIOU of $0.3$ and $0.5$ are equivalent to $D$ of $6.9$ m and $13.3$ m, respectively. These levels are both looser than our standard tolerance of $D=5$ m in the previous experiment.

Table \ref{tab:det_cmp} shows the comparison results. Our method outperforms MMInsectDet in all three metrics. Among them, our ACE is less than half (about $45$\%) of that of MMInsectDet, indicating higher accuracy. Moreover, the elevated precision implies that our method has a tighter error distribution and higher reliability. Regarding applicability, MMInsectDet requires intersection labels to train its deep learning network. This type of dataset is not currently available to the public and is laborious to create. In contrast, our method does not use intersection labels for training. Furthermore, the implementation of MMInsectDet needs to maintain a separate network to process the raw point cloud. Meanwhile, our method can leverage the existing road segmentation module in the vehicle. This improves flexibility and reduces the overall computational demand of the system.

\begin{figure} [t]
    \centering
    \includegraphics[width=\linewidth]{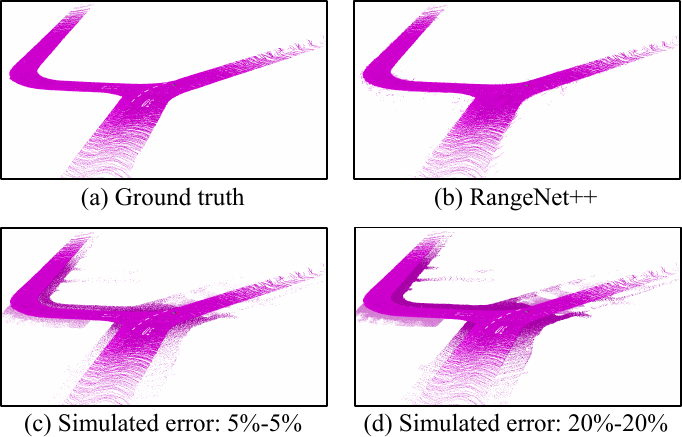}
    \caption{Different levels of road segmentation error at the same intersection. Magenta: true road points. Purple: false-positive road points.
    \vspace{-6mm} }
    \label{fig:seg_err}
\end{figure}

\subsection{Robustness to Road Segmentation Errors}

We tested the robustness of our method under challenging conditions of road segmentation errors that are higher than the benchmark in Section \ref{sec:performance}. In real-world operations, the semantic segmentation module can be less accurate due to weather, scenery, and sensor factors. This may affect our method, which uses road segmentation as an input. Through this experiment, we aim to investigate the degradation of our method with respect to the error level of this input.

\subsubsection{Segmentation Error Setup}

We use manual semantic labels from the SemanticKITTI dataset as the ground truth. We simulate different error conditions by randomly removing a fraction (r-FNR) of ground truth road points from each LiDAR scan, representing FNs. We also randomly use a certain proportion (r-FPR) of points from three other classes in the ground truth labels, including `sidewalk', `parking', and `other-ground', to simulate FPs. We use these three classes because they are most often confused with the `road' class. We generated four simulation variations of the r-FPR and r-FNR ratios: 5\%-5\%, 5\%-20\%, 20\%-5\%, and 20\%-20\%. Fig. \ref{fig:seg_err} illustrates the segmentation error conditions of the test. RangeNet++ road labels show minimal error relative to the ground truth, while the simulated cases display more noticeable inaccuracies.

\subsubsection{Robustness Analysis}

\begin{figure} [t]
    \centering
    \includegraphics[width=1\linewidth]{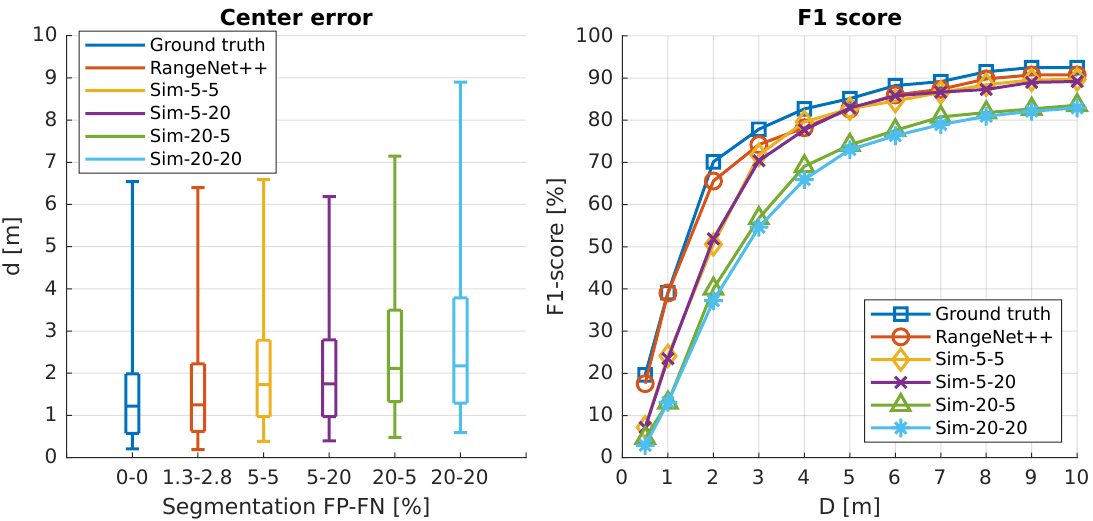}
    \caption{Performance of our method with different road segmentation error levels. Center error (left): each box plot represents the 5th, 25th, 50th, 75th, and 95th percentiles of the error distribution. F1-score (right): the score is computed at confidence thresholds from 0.5 m to 10 m.}
    \label{fig:segerrs_perf}
\end{figure}

Fig. \ref{fig:segerrs_perf} shows the performance variation of our method with different levels of segmentation error. For ease of presentation, we calculate the F1-score from precision and recall as: $\text{F1-score} = \frac{2\cdot \text{Pre} \cdot \text{Rec}}{\text{Pre}+\text{Rec}}$.
As the error rate increases from 0\%-0\% to 20\%-20\%, the center error increases by less than 1 m at the median value and less than 2.5 m at the 95\% distribution boundary. At the same time, the F1-score decreases by 12\% at the average confidence threshold of $D=5$ m. These indicate that our method has a low slope of performance degradation over segmentation errors.

\begin{table}[t]
    \centering
    \caption{Performance with different segmentation error levels}
    \begin{tabular}{l | c c | c c c} 
        \toprule
        Label source & r-FPR& r-FNR& ACE& Pre@5& Rec@5 \\
        \midrule
        Ground truth& 0.00 & 0.00 & 1.86 & 90.06 & 80.69\\
        RangeNet++& 1.34 & 2.84 & 1.92 & 89.48 & 76.74\\         
        \midrule        
        Sim-5-5 & 5.00 & 5.00 & 2.26 & 90.59 & 76.14\\
        Sim-5-20 & 5.00 & 20.00 & 2.32 & 90.59 & 76.44\\     
        Sim-20-5 & 20.00 & 5.00 & 2.94 & 80.95 & 68.39\\
        Sim-20-20 & 20.00 & 20.00 & 3.23 & 78.83 & 68.00\\      
        \bottomrule
    \end{tabular}
    \vspace{-5mm} 
    \label{tab:sem_err}
\end{table}

Table \ref{tab:sem_err} shows the quantitative results with different levels of segmentation errors. 
The segmentation from RangeNet++ is the benchmark level used in Section \ref{sec:performance}. With a low error rate of 1.3\%-2.8\%, the performance is close to the upper bound of using ground truth labels. This suggests that RangeNet++ is a reasonable choice to pair with our method. 
In the four simulated error cases, with the same amount of change, FNs have minimal effect on performance, while FPs have a greater effect. This shows that our method is more robust to FNs.
In cases with moderate error rates including 5\%-5\% and 5\%-20\%, the ACE only increases by 0.4 m, while precision and recall remain the same as the benchmark case. This demonstrates that our method is robust to FP and FN rates that are 3.7 and 7 times as high as those in the benchmark case, respectively.
In cases with significant error rates such as 20\%-5\% and 20\%-20\%, the results degrade noticeably. However, the ACE here is still 1 m smaller than that of MMInsectDet as shown in Table \ref{tab:det_cmp}, and the precision here is still around 80\%. This shows that our method still performs fairly in challenging cases.


\subsection{Limitations} \label{sec:exp_limit}

Our method faces challenges with compound intersections. These are most common in sequence 06. Compound intersections contain multiple intersection points clustered in a large common area. In such cases, the middle axis transform used in our method tends to produce corner points that are close together, causing subsequent processing to merge them into a single intersection point. This both increases the center error and decreases the number of detected points, significantly reducing precision and recall. 

Regarding the evaluation, the definition of the `road' class is not exactly the same in SemanticKITTI and OSM. Some paths are labeled as drivable roads in SemanticKITTI but not in OSM, and vice versa. Therefore, there are intersections identified in one data set, but not in the other. This leads to some uncertainty in the evaluation and may underestimate the detector more than it actually is. This phenomenon is most visible in sequences 09 and 10.

\section{Conclusion}\label{conclusion}

This paper presented a novel method for the detection and localization of road intersections that is based purely on vehicle-mounted 3D LiDAR. The method involves a six-step process that leverages the availability of road segmentation and vehicle's local pose estimate to extract the intersection candidates. This process is followed by a branch-based refinement model to classify and correct intersection points. In addition, the paper proposed a novel automated evaluation method that uses public georeferenced road maps instead of relying on manual labels. Experiments on eight real-world dataset sequences demonstrated that the proposed method outperforms a state-of-the-art method in accuracy and reliability. Tests with corrupted road segmentation input showed that the method has a high robustness to moderate segmentation errors and performs fairly in challenging cases. This robustness indicates that the method is highly applicable in practice. Future contributions can be made to the handling of compound intersections and the unification of the definition of the road class across datasets.

%








\bibliographystyle{ieeetr} 
\bibliography{references/paperref1} 

\end{document}